\documentclass[letterpaper]{article} 
\usepackage{aaai24}  
\usepackage{times}  
\usepackage{helvet}  
\usepackage{courier}  
\usepackage[hyphens]{url}  
\usepackage{graphicx} 
\usepackage{natbib}  
\usepackage{caption} 
\usepackage{algorithm}
\usepackage{algorithmic}

\usepackage{apalike}
\usepackage{cite}
\usepackage{textcomp}
\usepackage{makecell}
\usepackage{multirow,bigdelim,dcolumn,booktabs}
\usepackage{array}
\usepackage{float}
\usepackage{enumitem} 
\usepackage[caption=false]{subfig}
\usepackage{url}
\usepackage{xurl}
\usepackage{multirow, booktabs}
\usepackage{etoolbox}
\usepackage{threeparttable}
\usepackage{adjustbox}

\title{Graph Encoding and Neural Network Approaches for Volleyball Analytics: From Game Outcome to Individual Play Predictions}

\author{{Rhys Tracy}, {Haotian Xia}, {Alex Rasla}, {Yuan-Fang Wang}, {Ambuj Singh}}

\affiliations {Departement of Computer Science, University of California, Santa Barbara\\\{rhystracy, haotianxia, yfwang, ambuj\}@cs.ucsb.edu, alexrasla@engineering.ucsb.edu}

\begin{document}
\maketitle
\begin{abstract}
\begin{quote}
This research aims to improve the accuracy of complex volleyball predictions and provide more meaningful insights to coaches and players. We introduce a specialized graph encoding technique to add additional contact-by-contact volleyball context to an already available volleyball dataset without any additional data gathering. We demonstrate the potential benefits of using graph neural networks (GNNs) on this enriched dataset for three different volleyball prediction tasks: rally outcome prediction, set location prediction, and hit type prediction. We compare the performance of our graph-based models to baseline models and analyze the results to better understand the underlying relationships in a volleyball rally. Our results show that the use of GNNs with our graph encoding yields a much more advanced analysis of the data, which noticeably improves prediction results overall. We also show that these baseline tasks can be significantly improved with simple adjustments, such as removing blocked hits. Lastly, we demonstrate the importance of choosing a model architecture that will better extract the important information for a certain task. Overall, our study showcases the potential strengths and weaknesses of using graph encodings in sports data analytics and hopefully will inspire future improvements in machine learning strategies across sports and applications by using graph-based encodings.
\end{quote}
\end{abstract}

\section{Introduction}

As the sport of volleyball has gotten more popular and players have begun joining at younger and younger ages, the level of play has also increased. This increase in popularity and the subsequent improvements in player skill have lead to increasing demands for tactical analysis and better game strategies. These demands come with a greater necessity for computational analysis of the sport.

With recent increases in interest for sports data analytics through all sports, we have seen an increasing number of studies looking into predicting game events \cite{b31}, analyzing team and individual player performance\cite{b15}, overall sport development\cite{b3}, and predicting or analyzing overall team performance across sports. For example, the sports of Basketball, Soccer, and Baseball have seen several studies.

Basketball has seen several datasets released and other analysis\cite{b24,b26}, (Miljković et al. 2010)\cite{b27,b32,b33} for predicting game outcomes, improving player developement, predicting rising stars, or identifying opposing team's offensive and defensive strategies. Soccer \cite{b31,b11,b23,b25,b28,b4,b5,b6,b7,b8,b9,b10} have also seen studies focused toward a variety tasks--including game event and outcome predictions, posture analysis, game lineup prediction, and injury risk assessment.

Despite growing interest in sports analytics, studies on the sport of Volleyball have been limited so far. The few studies that have been conducted have tended to have small scopes and use basic naive approaches, yet they have yielded a promising start and a good baseline to compare with. With the increasing need for more and more sophisticated data analytics strategies, we wish to introduce specialized encodings and models for the sport of volleyball to improve upon these current baseline approaches without the need for gathering additional data.

\section{Related Work}

There have been a few recent datasets and studies for indoor volleyball\cite{b13,b21}, but they are primarily focused toward computer vision and are not the most useful for tactical analysis or tracking in-depth game statistics since they are missing several important game variables. There has also been a recent beach volleyball dataset\cite{b21} that has been more useful for tactical analysis, but due to differences between beach volleyball and indoor volleyball--primarily the additional players and more strict positions in indoor volleyball--this dataset is limited solely to beach volleyball analysis. Lastly, a recent study\cite{b34} introducing a specialized indoor volleyball dataset has made a noticeable leap in the field.

Xia et al.\cite{b34} has introduced a simple yet powerful natural language to represent a volleyball rally as a sequence of rounds that each consist of a sequence of 1-4 contacts (pass, set, hit, block) and gathered a large dataset of NCAA and Professional men's volleyball games. This dataset has allowed much deeper and more useful game statistics to be captured and analyzed, and this paper also demonstrated some promising results in several never-before-analyzed tactical analysis tasks. The authors demonstrated some impressive results with simple naive models when predicting the winner of a rally, predicting where a setter will set the ball, and predicting what type of hit a hitter will use. All of these tasks are valuable for a team to determine more optimal offensive and defensive strategies under different scenarios. This paper does, however, leave room for improvement. The data used in this study is raw, the approaches are naive, and the authors never explore any forms of encoding their data to improve performance since they are simply introducing baselines that their language is capable of. Given the temporal sequence heavy format of the volleyball language, we decided to explore temporal graph based encodings for this language and dataset.

Graph based encodings have shown promising results for analyzing other sports, such as American football, basketball, and soccer. One recent study uses graph encodings and Graph Neural Networks to predict various sports outcomes\cite{b35}. This paper focuses on creating a sports-agnostic way of representing game-states using graph encodings. Through this technique, they are able to capture inter-player relationships and local player relationships that can otherwise not be taken into account when training a model. Further, the paper tests its player-specific graph approach on American football and a popular esports game, Counter-Strike. Through their methods, they demonstrate a reduction in test loss by 20\% and 9\% for football and Counter-Strike respectively. A similar work uses GNNs for predicting future player locations and movements\cite{b36}. The work focuses on multi-agent sports such as basketball and soccer and takes advantage of both GNNs and Variational RNNs to generate these future locations. Through these techniques, they show that the statistical player distribution of their generative model predictions outperform previous works. Further, they also test their model using conditional prediction to answer questions such as: How will the players move if A passes to B instead of C? Insight into these conditional relationships is highly tactically useful when determining new strategies or how to optimize individual play.

Since many sports can have data be well represented with graph structures, graph encodings would likely work well in the sport of volleyball in several areas. With the current leading dataset\cite{b34} especially, using a graph encoding might better represent the data. For instance, the original dataset does not give the deep learning models any information as to which variables belong to which contact in a given "ball round". On top of this, there is no information describing the temporal ordering of the contacts given to the models. To give this additional context information to the deep learning models and allow for a better representation of a volleyball rally without needing to gather any additional data, we propose encoding the volleyball round data into a graph structure and using Graph Neural Networks to complete the same tasks.

\section{Graph Encoding}

To analyze our graph encoding, we must first look at what information the baseline dataset has to offer.

 \begin{figure*}[htbp]
 \centering
\includegraphics[width=1\textwidth]{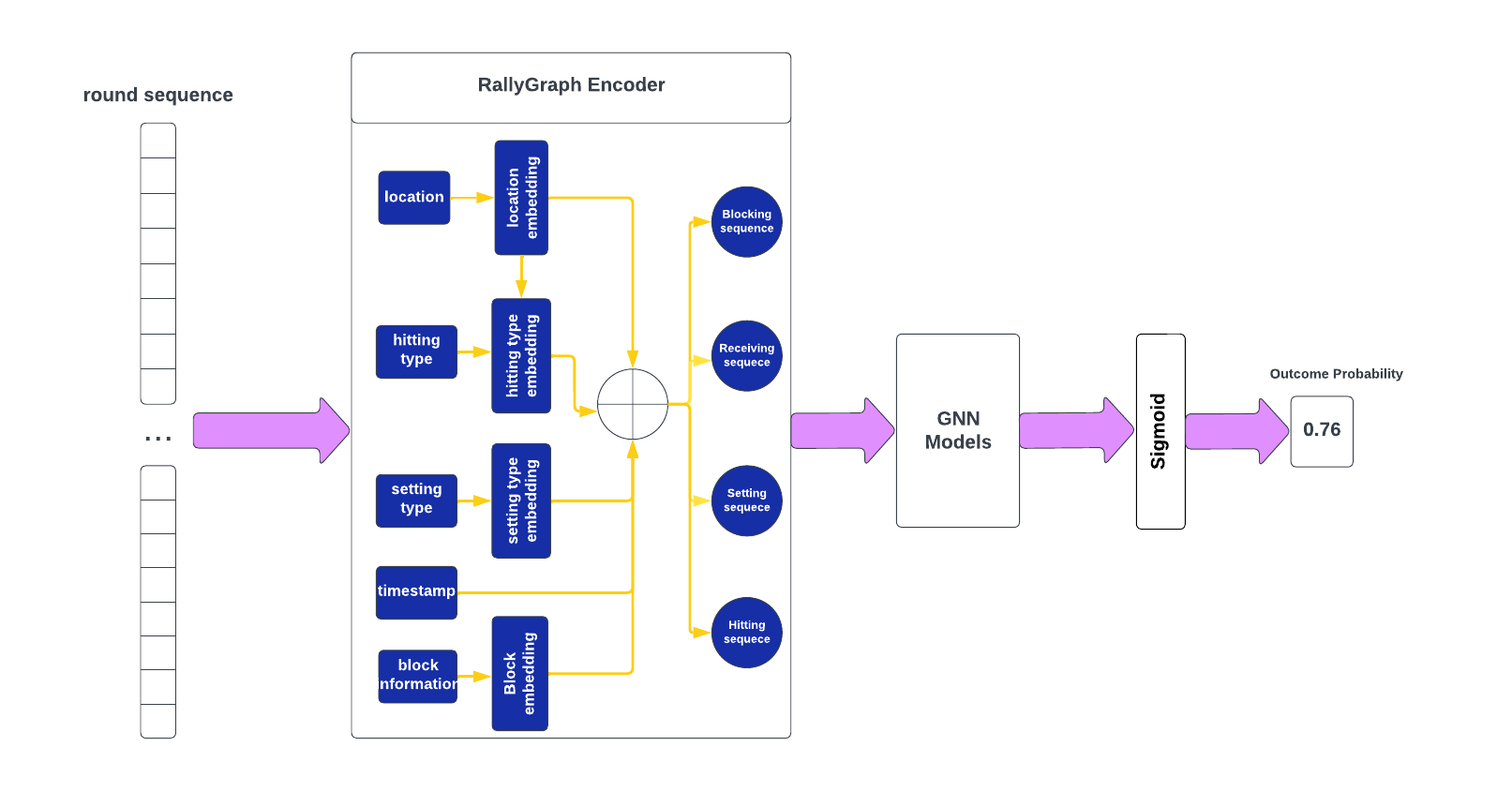}
\caption{Our framework for taking raw round sequence data, encoding into a Graph structure, then making a rally outcome prediction}
\label{mdoel}
\end{figure*}

\subsection{Underlying Data Representation}

For this study, we will use the leading indoor volleyball dataset (Xia et al. 2022)\cite{b34}. This dataset splits volleyball matches into a sequence of rallies and splits each rally into a sequence of rounds. Each round contains variables describing round information (team and round number), various locations of ball contacts, pass and set ratings, hit type used, blocking information, and serve type. As explained earlier, all of this information---besides the two round level variables---relates to each individual ball contact. Pass contact location and pass rating relate to the pass contact, set rating and location relate to the set contact, etc. All of this points to a contact-level encoding being an excellent option to test. Traditional sports GNN based approaches usually focus on mapping player interactions with graph structures as the interactions and locations of players are typically the most important variables to analyze for free-flowing sports such as basketball or soccer, however since the sport of volleyball has the teams separated by the net and a very rigid rule set to follow, the player interactions and locations become less useful to analyze since there is less freedom and instead the interactions with and locations of the ball and ball contacts become much more important information to focus on. Thus, this VREN Dataset\cite{b34} primarily focuses on ball interactions, and our analysis with a contact-level graph encoding will attempt to augment the information from these ball interactions.

\subsection{Encoding Methods}

In order to attach the variables to their correct contacts and encode the temporal format of the contacts, we must consider each contact as a node of the graph and add each contact's important information to that node. For example the set contact node will hold the setter location, set rating, and set destination variables and the hit contact node will hold the hitter location and hit type variables. Then to encode the temporal aspect of the contacts, we will connect each consecutive contact with a one way edge from first contact to second contact. For example, in a round with a pass, a set, and a hit, we would connect the nodes with one way edges from the pass node to the set node and from the set node to the hit node. All edges will have equal weights, and the dataset does not include any other useful information to include in edge attributes. Though a simple graph encoding method, it fundamentally changes how a neural network will analyze the data.

\subsection{Rally Outcome Prediction Task}

Since the baseline rally outcome prediction task in the VREN paper\cite{b34} considers all information in a round (except the winning team and win/lose reason), we will use all the nodes for a given round to make a prediction. As such, we end up with a graph involving a pass node, then set node, then hit node, then block node. These graphs will involve the exact same information as is used in the baseline task, but will just have a new graph encoding.

\subsection{Set Location Prediction Task}

The baseline set location prediction task in the VREN paper only uses the information in a round up to when the setter is about to set the ball. If we were to follow this same strategy, our encoding would involve only 2 nodes (pass and set nodes) in each graph, and very small graph sizes seem to yield poor GNN performance. However, we can include information from the hit and block nodes of the previous round (if there is a previous round in that rally). From further baseline testing, this additional information does not noticeably effect performance in any baseline model, so any performance changes with this task will be solely from the graph encoding. Therefore, for this task each graph involves the previous round's hit node (if it exists), the previous round's block node (if it exists), the current round's pass node, and the current round's set node only including information from before the setter contacts the ball---such as where the setter will set the ball from.

\subsection{Hit Type Prediction Task}

Similar to the set location prediction task, the baseline hit type prediction task in the VREN paper only uses the information in a round up to when a hitter is about to hit the ball. For this task, we decided to add in the previous round's block node to reach a consistent graph size of 4 nodes for better GNN performance. As with the set prediction task, this additional information yielded no difference in baseline performance. So for this task, each graph involves the previous round's block node (if it exists), the current round's pass node, the current round's set node, and the current round's hit node only including information from before the hitter contacts the ball---such as the location the hitter will be hitting from.

\section{Methods}

With the graph encodings set up, next we turn to the models we will test. To keep comparison consistent with the baseline, we will test a GCN to compare with CNN, a Graph GRU to compare with LSTM, and a Graph Transformer to compare with Transformer. To Implement all of these models, we will use Spektral, a GNNs package built on Tensorflow and Keras. Since this package does not include a Graph Transformer Convolution layer, we implemented one modeled off of the Graph Transformer architecture introduced in Shi et al.\cite{b37}  which has shown excellent results for Graph-based learning tasks. We used Tensorflow and Keras to build this Graph Transformer Convolution Layer off of the base MessagePassing layer available in Spektral. This Graph Transformer architecture performs self-attention on graph edges with queries embedded from the node features for the origin of the edge and keys and values embedded from the node features for the terminal of the edge. This architecture also includes gated residual connections between layers, a key factor making this architecture a Transformer.

\subsection{Models Tested}

For the rally outcome prediction task, we will analyze all three of these GNN structure's performances just like the baseline, but for the other two tasks, we will analyze the Transformer and CNN/GCN architectures only since the RNNs tend to struggle on these volleyball tasks for a number of reasons. The GCN architecture we used for all tasks involved one Graph Convolution Layer, a graph global pooling layer to get a graph level value, then 3 dense layers. The Graph GRU architecture we used involved a Gated Graph Convolution Layer with a GRU Gate, a graph global pooling layer, then 2 dense layers. Lastly the Graph Transformer architecture we used for all tasks invloved one custom Graph Transformer Convolution Layer, a graph global pooling layer, then 2 dense layers. The rally outcome models had a single float output with a sigmoid activation function on the final output layer to yield a probability value between 0 and 1; these models used MSE as their loss function for training, and MAE, binary accuracy, AUC, and Brier Score as metrics for validation and testing. The set location models had an array of 9 values as an output with a softmax activation function on the final output layer to yield probability predictions for each set location that sum to 1; these models used the cross-entropy loss function for training and categorical accuracy as metrics for validation and testing. The hit type models had an array of 8 values as an output with a softmax activation function on the final output layer to yield probability predictions that sum to 1; these models also used the cross-entropy loss function for training and categorical accuracy as metrics for validation and testing.

\subsection{Hit Type Prediction Task Modification}

We decided to modify the original Hit Type Prediction task to make it even more useful for tactical analysis. To do this, we decided to ignore blocked hits. First, this will make it easier to predict hit types as the other hit types are relatively predictable given the set rating, locations, and other information in this dataset, but whether the ball is blocked or not depends on much more than the information in the dataset. This means that including the "blocked" hit type will end up confusing the models since they are missing the necessary information to predict if a player will be blocked. Second, removing blocked hits will focus the task on more important predictions. Predicting if a hitter will be blocked provides essentially no use in tactical analysis, as strategy on both the offensive and defensive sides will not change if you know that a hitter will or will not be blocked. The offensive team will see no tactical gain in not covering their hitter when they know they won't be blocked and the defensive team will see no tactical gain in allowing their non-blocking players to relax on the court because they know the ball will be blocked. The other hitting types are much more tactically useful to be able to predict, so we will focus the Hit Prediction Task on those hitting types and seek to improve performance there. We ran the same baseline models on our modified version of this task for comparison with our graph-based approach.

\section{Results and Analysis}

We found that the graph encodings provided noticeable improvements in most cases and consistently standardized models' performances between the NCAA and Professional games. In the baseline models, the models would consistently perform much better or worse on either the NCAA or Pro game for every task. This was due to differences in the level of play and consistency between the NCAA and Professional games in the dataset. For example, the baseline models were better at predicting rally outcomes and hit types in the professional games because they are less random and more deterministic given the better, more consistent, and more mentally strong play by professional players; additionally the baseline models were better at predicting the set locations because professional setters are more skilled and make more randomized sets to confuse the opposing team. With the graph encodings, however, the difference in model performance between NCAA and Pro games was small or negligible in almost all cases. This decrease in this performance gap is because the graph encoding makes the underlying volleyball relationships between the data much more clear to the models. In some tasks, these relationships are more or less clear to the baseline models in NCAA or Professional play, but by manually explaining these relationships to the models, the relationships become equally clear and thus the models' performances becomes more similar. For most cases, this improved understanding of the underlying relationships between contacts in the game can boost performance, and for others---where contact-by-contact information may not be as useful or where models had difficulty analyzing the additional information---it struggled to improve performance.

Next we present an in depth analysis of each task's results.

\subsection{Rally Outcome Prediction Results}

 \begin{table*}[h]
\centering
\caption{rally outcome prediction task performance of each model on college-level games \& professional games. There are significant improvements among all three models compared to the baseline performance. The Graph Transformer gives the best result.}
\label{hitting prediction}
\begin{adjustbox}{totalheight=0.25\textheight}
\begin{threeparttable}
\begin{tabular}{cccccc}
\Xhline{1.2pt}
\makecell{Level of game}&\makecell{Model} &\makecell{Binary Accuracy(\%)}&\makecell{AUC}&\makecell{Brier Score}&\makecell{Mean Absolute Error} \\
 \hline
 \multirow{6}{*}{\makecell{college}} 
  & Transformer$^*$ & 74.38& 0.82&0.18 &0.34\\
  & CNN$^*$ & 69.06 & 0.75&0.20 &0.40\\
  & LSTM$^*$ & 65.91& 0.75&0.21 &0.41\\
  & Graph Transformer & \textbf{81.15}& \textbf{0.87}& \textbf{0.15} &\textbf{0.27}\\
  & Graph GRU & 77.81 & 0.86 & 0.33 & 0.31\\
  & GCN & 78.20 & 0.86&0.20 &0.30\\

  \hline
 \multirow{6}{*}{\makecell{professional}}
  & Transformer$^*$  & 80.00& 0.85& \textbf{0.16} &0.32\\
  & CNN$^*$  & 71.59 & 0.76& 0.20 &0.39\\
  & LSTM$^*$  & 70.06& 0.75& 0.20 &0.40\\
  & Graph Transformer & \textbf{81.15}& \textbf{0.87}& \textbf{0.16} &\textbf{0.27}\\
  & Graph GRU & 77.83 & 0.86 & 0.29 & 0.31\\
  & GCN & 78.20 & 0.86& 0.17 &0.30\\

\Xhline{1.2pt}
\end{tabular}
\begin{tablenotes}
          \footnotesize  
          \item[*] model prediction results from VREN\cite{b34}.
        \end{tablenotes}
\end{threeparttable}
\end{adjustbox}
\end{table*}

 Overall, the use of our graph encodings and GNNs yielded significant improvements for the rally outcome prediction task as shown in Table~\ref{hitting prediction} below. GCN and Graph GRU both yielded huge improvements for both the NCAA and Pro testing games over baseline in all metrics but brier score. Graph Transformer yielded a large improvement in NCAA game performance and a slight but noticeable boost in Pro game performance in all metrics except for brier score (which performed slightly better in the Pro game and slightly worse in the NCAA game). These improvements suggest that the graph encoding gives the model a much more detailed picture of what is happening in the rally allowing for more detailed---and thus better performing---predictions. Since a Transformer does an excellent job at analyzing the relationships between different variables with its attention mechanism, it does not benefit as much from the graph encoding. The Graph Transformer did perform noticeably better on the NCAA game, however, so this again suggests that the lower level of play in the NCAA game made it harder to find these underlying relationships as compared to the Pro game, but the graph encoding was able to make these relationships more clear and thus standardized performance between the two levels of play.

 \subsection{Set Location Prediction Results}
 
 \begin{table}[h]
\centering
\caption{Categorical Accuracy for setting location prediction in both professional and college level games. All three models are improved compared to the baseline result}
\begin{center}
\label{Set Prediction}
\begin{adjustbox}{width=\columnwidth}
\begin{threeparttable}
\begin{tabular}{ccc}
\Xhline{1.5pt}
\makecell{Level of game}&\makecell{Model} &\makecell{ Categorical Accuracy(\%)} \\
 \hline
 \multirow{4}{*}{\makecell{college}} & Transformer$^*$& 54.65\\
    & GCN & \textbf{59.10} \\ 
    & CNN & 57.43 \\
    & Graph Transformer & 56.57\\
  \hline
 \multirow{4}{*}{\makecell{professional}} & Transformer$^*$& 51.65\\
  & GCN & \textbf{59.10}\\
  & CNN & 53.30 \\
  & Graph Transformer & 56.57\\
 
\Xhline{1.5pt}
\end{tabular}
\begin{tablenotes}
          \footnotesize  
          \item[*] model prediction results from VREN\cite{b34}.
        \end{tablenotes}
\end{threeparttable}
\end{adjustbox}
\end{center}
\end{table}

 The set location prediction task saw noticeable improvements when using our graph encoding as shown in Table~\ref{Set Prediction}. Not only did a base CNN perform better than the baseline Transformer (the only model tested in VREN\cite{b34}), but the GCN improved upon the CNN's performance even further. Additionally the Graph Transformer saw a similar performance boost over the baseline Transformer. With our addition of the baseline CNN and the improved performance with the graph encodings, we found a much better understanding of what information goes into a setters decision to set the ball. First, given the fact that setters try to be as random as they can be when making set location predictions, a simpler model seems to do better. The simple CNN model gives a 2-3\% performance boost over the baseline Transformer, which is pretty noticeable in a hard to predict task, and GCN yields a similar performance boost over Graph Transformer. Secondly, it seems that the graph encoding and additional information from the previous round does help improve set prediction performance, but the improvement is not as significant as the previous task. This result, combined with the better performance of the convolution models over the transformer models, would suggest that the location a setter will set to depends less on contact-by-contact information than the outcome of the rally. Instead, the location a setter will set to likely depends more on simpler information (such as setters location on the court, pass rating, etc), which would allow a simpler convolution model to extract this information better. Additionally, this theory would match up with volleyball expert's opinions on what influences the set location during a rally. Lastly, there is likely other useful information for predicting where a setter will set the ball that is not included in this dataset, such as how rushed the setter is and how high the pass is.

\subsection{Hit Type Prediction Results}

\begin{table}[h]
\centering
\caption{Categorical Accuracy for hit type prediction task in both NCAA and Professional games with blocked hits included vs excluded from training and testing. Graph Transformer improved on or gave consistent results with baseline in all cases, while GCN performed slightly worse than baseline in all cases}
\label{res}
\begin{adjustbox}{width=\columnwidth}
\begin{threeparttable}
\begin{tabular}{cccc}
\Xhline{1.5pt}
\makecell{Level of game}&\makecell{Blocked hits}&\makecell{Model} &\makecell{Categorical accuracy(\%)}\\[0.5mm]
 \hline \\
 \multirow{10}{*}{\makecell{College}} & \multirow{4}{*}{included} 
  & Transformer$^*$& 71.28\\
  & & GCN & 69.64\\
   & & CNN & 72.04\\
   & & Graph Transformer & \textbf{73.31}\\[1.5mm]\cline{2-4}
 \\& \multirow{4}{*}{excluded} 
 & Transformer$^*$& 80.68\\
  & & GCN & 80.10\\
   & & CNN & 80.68\\
   & & Graph Transformer & \textbf{86.41}\\
   \\\hline
 \\\multirow{10}{*}{\makecell{Professional}} & \multirow{4}{*}{included} 
 & Transformer$^*$& 73.63\\
  & & GCN & 69.64  \\
  & & CNN & \textbf{74.73}\\
   & & Graph Transformer & 73.31\\[1.5mm]\cline{2-4}\\
& \multirow{4}{*}{excluded} 
& Transformer$^*$& 86.36\\
  & & GCN & 80.10\\
   & & CNN & 86.36\\
   & & Graph Transformer & \textbf{86.39}\\[1.5mm]
 
\Xhline{1.5pt}
\end{tabular}
\begin{tablenotes}
          \footnotesize  
          \item[*] model prediction results from VREN\cite{b34}.
        \end{tablenotes}
\end{threeparttable}
\end{adjustbox}
\end{table}

 For the hit prediction task, the graph encodings and additional information from the previous round yielded mixed results but for the most part not much change as shown in Table~\ref{res}. The graph encoding noticeably improved the performance of the Graph Transformer over the baseline for the NCAA games and was relatively the same for the Pro games, and the GCN performed slightly worse than baseline CNN for both sets of games. These results would suggest that the encoded contact-by-contact information is not as important for this task; the additional information may slightly improve models that can use this information well---such as a Transformer---but may harm simpler models that cannot analyze this information as well. Overall this would point to the location a hitter hits to being heavily influenced by individual variables, but with some small influence from other contact-by-contact information. When including the "blocked" hit type, the simpler base CNN model outperformed the baseline Transformer model, but when this hit type is excluded they performed identically. This would suggest that a player getting blocked may depend more heavily on simple information (such as the location of the hitter or number of blockers), but the hit type a hitter chooses to use may depend slightly more on the contact-by-contact information. Additionally, removing the "blocked" hit type yielded significant improvements in performance in all cases. Just like the set prediction task, this task would very likely benefit from having additional information, such as how tight to the net the set is and where the ball is located in relation to the hitter's body.

 \subsection{Analysis}

 From these results, we can gather that predictions for outcomes that are largely influenced by contact-level information (like the rally outcome) will largely benefit from using a graph encoding to bolster the contact-level information inputted into a model. And similarly, outcomes that depend on contact-level information less (like the type of hit a hitter uses) will benefit less or see mixed results. Additionally, the use of our graph encoding allows the models to analyze the game in a much more advanced and human way. The baseline models view the variables in the round globally and then they try to analyze the connections between them. This is similar to how a beginner of the sport would view a volleyball game: they view a whole round and understand general connections between big events that factor into certain outcome---like which team wins the rally. To truly understand the flow of a rally, however, it’s more important to look at how consecutive contacts are influencing each other. For example, it’s more important to understand how serving to a certain player may affect the rating and location of the pass, which may limit the players a setter can set to, which can significantly decrease a teams chance of winning that rally. This sort of analysis is already used by coaches, analysts, and fans who are more knowledgeable of the sport to evaluate how a game is going and how to change strategy to improve. GNNs primarily focus their analysis on graph edges, and in our encoding specifically this is the relationships between consecutive contacts. Thus the use of GNNs with our graph encodings yields a much more advanced analysis of the data which noticeably improved our results overall. Another takeaway is that these baseline tasks can be significantly improved---both in prediction ability and in tactical usefulness---with simple adjustments. Removing blocked hits greatly improved prediction results at the same time as making the predictions more tactically focused and applicable.

\section{Conclusion and Future Work}

In this paper, we introduce a novel graph encoding to add additional contact-by-contact volleyball context to an already available volleyball dataset without any additional data gathering. This graph encoding can yield large improvements in prediction tasks that depend heavily on this information, but may not yield much benefit for tasks that do not. Ultimately, encodings are specialized tools that will not work in all situations. Overall these results show the potential strengths and weaknesses of using graph encodings in sports data analytics and hopefully will inspire future improvements in machine learning strategies across sports and applications by using graph based encodings. Additionally, we demonstrate the importance of choosing a model architecture that will better extract the important information for a certain task---in some sports analysis tasks a much simpler model will perform noticeably better on the given data. Lastly, we were also able to gain a much better understanding of the underlying relationships in a volleyball rally from these results; for a coach or player, this is extremely useful for making more informed game decisions.

In future studies, we hope to gather more sophisticated data than included in the dataset we explored in this study and analyze other encoding formats to see if that can improve prediction results.
\bibliography{aaai24}

\begin{thebibliography}{24}
\providecommand{\natexlab}[1]{#1}

\bibitem[{Aoki(2010)}]{b10}
Aoki, K. 2010.
\newblock Plays from Motions for Baseball Video Retrieval.
\newblock 271--275.

\bibitem[{Baboota and Kaur(2019)}]{b25}
Baboota, R.; and Kaur, H. 2019.
\newblock Predictive analysis and modelling football results using machine
  learning approach for English Premier League.
\newblock \emph{International Journal of Forecasting}, 35(2): 741--755.

\bibitem[{Chun, Son, and Choo(2021)}]{b9}
Chun, S.; Son, C.-H.; and Choo, H. 2021.
\newblock Inter-dependent LSTM: Baseball Game Prediction with Starting and
  Finishing Lineups.
\newblock 1--4.

\bibitem[{Claudino et~al.(2019)Claudino, de~Oliveira~Capanema, de~Souza,
  Serrão, Pereira, and Nassis}]{b15}
Claudino, J.~G.; de~Oliveira~Capanema, D.; de~Souza, T.~V.; Serrão, J.~C.;
  Pereira, A. C.~M.; and Nassis, G.~P. 2019.
\newblock Current approaches to the use of artificial intelligence for injury
  risk assessment and performance prediction in team sports: a systematic
  review.
\newblock \emph{Sports medicine-open}, 5(1): 1--12.

\bibitem[{Decroos et~al.(2019)Decroos, Bransen, Haaren, , and Davis}]{b11}
Decroos, T.; Bransen, L.; Haaren, J.~V.; ; and Davis, J. 2019.
\newblock Actions Speak Louder Than Goals: Valuing Player Actions in Soccer.
\newblock 1851–1861.

\bibitem[{Huang and Li(2021)}]{b7}
Huang, M.-L.; and Li, Y.-Z. 2021.
\newblock Use of Machine Learning and Deep Learning to Predict the Outcomes of
  Major League Baseball Matches.
\newblock \emph{Applied Sciences}, 11(10): 4499.

\bibitem[{Ibrahim et~al.(2016)Ibrahim, Muralidharan, Deng, Vahdat, and
  Mori}]{b13}
Ibrahim, M.; Muralidharan, S.; Deng, Z.; Vahdat, A.; and Mori, G. 2016.
\newblock A Hierarchical Deep Temporal Model for Group Activity Recognition.
\newblock 1971--1980.

\bibitem[{Jain and Kaur(2017)}]{b24}
Jain, S.; and Kaur, H. 2017.
\newblock Machine learning approaches to predict basketball game outcome.
\newblock 1--7.

\bibitem[{Mahmood, Daud, and Abbasi(2021)}]{b32}
Mahmood, Z.; Daud, A.; and Abbasi, R.~A. 2021.
\newblock Using machine learning techniques for rising star prediction in
  basketball.
\newblock \emph{Knowledge-Based Systems}, 211.

\bibitem[{McPherson and MacMahon(2008)}]{b8}
McPherson, S.; and MacMahon, C. 2008.
\newblock How Baseball Players Prepare to Bat: Tactical Knowledge as a Mediator
  of Expert Performance in Baseball.
\newblock \emph{Journal of Sport and Exercise Psychology}, 30(6): 755--778.

\bibitem[{Miljković et~al.(2010)Miljković, Gajić, Kovačević, and
  Konjović}]{b27}
Miljković, D.; Gajić, L.; Kovačević, A.; and Konjović, Z. 2010.
\newblock The use of data mining for basketball matches outcomes prediction.
\newblock 309--312.

\bibitem[{Nadikattu(2020)}]{b3}
Nadikattu, R.~R. 2020.
\newblock Implementation of new ways of artificial intelligence in sports.
\newblock \emph{Journal of Xidian University}, 14(5): 5983--5997.

\bibitem[{Prasetio and Harlili(2016)}]{b28}
Prasetio, D.; and Harlili, D. 2016.
\newblock Predicting football match results with logistic regression.
\newblock 1--5.

\bibitem[{Rudrapal et~al.(2020)Rudrapal, Boro, Srivastava, and Singh}]{b23}
Rudrapal, D.; Boro, S.; Srivastava, J.; and Singh, S. 2020.
\newblock A deep learning approach to predict football match result.
\newblock 93--99.

\bibitem[{Sawchik(2015)}]{b4}
Sawchik, T. 2015.
\newblock \emph{Big data baseball: Math, miracles, and the end of a 20-year
  losing streak}.
\newblock Macmillan.

\bibitem[{Shi et~al.(2021)Shi, Huang, Feng, Zhong, Wang, and Sun}]{b37}
Shi, Y.; Huang, Z.; Feng, S.; Zhong, H.; Wang, W.; and Sun, Y. 2021.
\newblock Masked Label Prediction: Unified Message Passing Model for
  Semi-Supervised Classification.
\newblock arXiv:2009.03509v5.

\bibitem[{Simpson et~al.(2022)Simpson, Beal, Locke, and Norman}]{b31}
Simpson, I.; Beal, R.~J.; Locke, D.; and Norman, T.~J. 2022.
\newblock Seq2Event: Learning the Language of Soccer Using Transformer-based
  Match Event Prediction.
\newblock 3898–3908.

\bibitem[{Sun, Lin, and Tsai(2022)}]{b6}
Sun, H.-C.; Lin, T.-Y.; and Tsai, Y.-L. 2022.
\newblock Performance prediction in major league baseball by long short-term
  memory networks.
\newblock \emph{International Journal of Data Science and Analytics}, 1--12.

\bibitem[{Thabtah, Zhang, and Abdelhamid(2019)}]{b26}
Thabtah, F.; Zhang, L.; and Abdelhamid, N. 2019.
\newblock NBA game result prediction using feature analysis and machine
  learning.
\newblock \emph{Annals of Data Science}, 6(1): 103--116.

\bibitem[{Tian et~al.(2019)Tian, Silva, Caine, and Swanson}]{b33}
Tian, C.; Silva, V.~D.; Caine, M.; and Swanson, S. 2019.
\newblock Use of Machine Learning to Automate the Identification of Basketball
  Strategies Using Whole Team Player Tracking Data.
\newblock \emph{Applied Sciences}, 10(1): 24.

\bibitem[{Wenninger, Link, and Lames(2020)}]{b21}
Wenninger, S.; Link, D.; and Lames, M. 2020.
\newblock Performance of machine learning models in application to beach
  volleyball data.
\newblock \emph{Int. J. Comput. Sci. Sport}, 19.

\bibitem[{Whiteley(2007)}]{b5}
Whiteley, R. 2007.
\newblock Baseball throwing mechanics as they relate to pathology and
  performance - a review.
\newblock \emph{Journal of sports science \& medicine}, 6(1): 1--20.

\bibitem[{Xenopoulos and Silva(2021)}]{b35}
Xenopoulos, P.; and Silva, C. 2021.
\newblock Graph Neural Networks to Predict Sports Outcomes: A Study on NCAA
  March Madness.
\newblock 35(18): 15757–15765.

\bibitem[{Xia et~al.(2022)Xia, Tracy, Zhao, Fraisse, Wang, and Petzold}]{b34}
Xia, H.; Tracy, R.; Zhao, Y.; Fraisse, E.; Wang, Y.-F.; and Petzold, L. 2022.
\newblock VREN: Volleyball Rally Dataset with Expression Notation Language.
\newblock 337--346.

\end{thebibliography}

\end{document}